\pdfoutput=1
\documentclass[11pt]{article}
\usepackage[final]{acl}
\usepackage{amsmath}
\usepackage{times}
\usepackage{latexsym}
\usepackage{bm}
\usepackage{booktabs}
\usepackage{makecell}
\usepackage{latexsym}
\usepackage{amsmath}
\usepackage{amssymb}
\usepackage{microtype}
\usepackage{inconsolata}
\usepackage{soul}
\usepackage{url}
\usepackage{graphicx}
\usepackage{amsthm}
\usepackage{booktabs}
\usepackage{algorithm}
\usepackage{algorithmic}
\usepackage[switch]{lineno}
\usepackage{subfig}
\usepackage{amsfonts}
\usepackage[T1]{fontenc}
\usepackage[utf8]{inputenc}
\usepackage{microtype}
\usepackage{inconsolata}
\usepackage{graphicx}
\usepackage{authblk}

\title{Recurrent Alignment with Hard Attention for Hierarchical Text Rating}

\author{%
Chenxi Lin\textsuperscript{1,2}, Jiayu Ren\textsuperscript{1}, Guoxiu He\textsuperscript{1,3}\thanks{Corresponding author.}, Zhuoren Jiang\textsuperscript{2}, Haiyan Yu\textsuperscript{1} and Xiaomin Zhu\textsuperscript{4} \\
\textsuperscript{1}School of Economics and Management, East China Normal University \\
\textsuperscript{2}School of Public Affairs, Zhejiang University \\
\textsuperscript{3}National Experiment Base for Intelligent Evaluation and Governance, Fudan University \\
\textsuperscript{4}Strategic Assessments and Consultation Institute, AMS \\
\texttt{\{cxlin,jyren\}@stu.ecnu.edu.cn,
gxhe@fem.ecnu.edu.cn, jiangzhuoren@zju.edu.cn,} \\
\texttt{hywei@infor.ecnu.edu.cn, xmzhu@nudt.edu.cn} 
}

\begin{document}

\maketitle
\begin{abstract}
While large language models (LLMs) excel at understanding and generating plain text, they are not tailored to handle hierarchical text structures or directly predict task-specific properties such as text rating. In fact, selectively and repeatedly grasping the hierarchical structure of large-scale text is pivotal for deciphering its essence. To this end, we propose a novel framework for hierarchical text rating utilizing LLMs, which incorporates \textbf{R}ecurrent \textbf{A}lignment with \textbf{H}ard \textbf{A}ttention (\textbf{RAHA}). Particularly, hard attention mechanism prompts a frozen LLM to selectively focus on pertinent leaf texts associated with the root text and generate symbolic representations of their relationships. Inspired by the gradual stabilization of the Markov Chain, recurrent alignment strategy involves feeding predicted ratings iteratively back into the prompts of another trainable LLM, aligning it to progressively approximate the desired target. Experimental results demonstrate that RAHA outperforms existing state-of-the-art methods on three hierarchical text rating datasets. Theoretical and empirical analysis confirms RAHA’s ability to gradually converge towards the underlying target through multiple inferences. Additional experiments on plain text rating datasets verify the effectiveness of this Markov-like alignment. Our data and code can be available in \url{https://github.com/ECNU-Text-Computing/Markov-LLM}. 
\end{abstract}

\section{Introduction}
\label{sec:intro}

Scaling up LLMs yields significant advances in their ability to mimic human-like text comprehension and generation \cite{ouyang2022training,zeng2022glm,touvron2023llama,openai2023gpt}. They demonstrate remarkable aptitude for in-context learning (ICL) \cite{brown2020language,min2022rethinking,kojima2022large} across various natural language processing (NLP) tasks \cite{qi2023art,chen2023personalized,wen2023unveiling,du2023task}. In particular, employing chain of thought (CoT) prompts can stimulate the reasoning capabilities of LLMs, enabling them to adeptly navigate and conquer complex downstream tasks \cite{wei2022chain,wang2023plan}.

\begin{figure}
\centering
\includegraphics[width=\columnwidth]{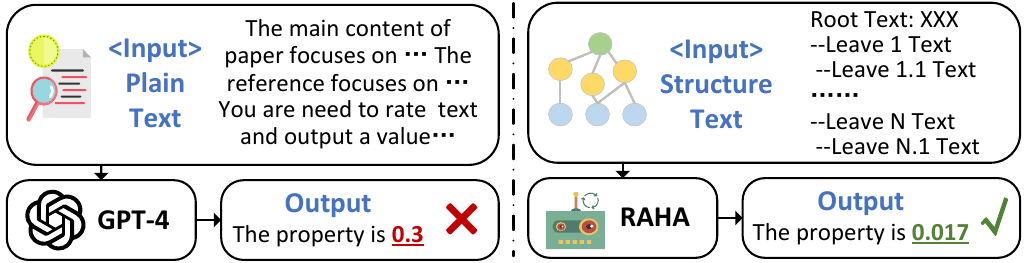}
\caption{A comparison between a typical LLM and our RAHA in processing hierarchical text rating task. While a typical LLM treats the input as plain text, our RAHA captures hierarchical structures and can straightforwardly provide task-specific rating score.}
\label{fig:task}
\end{figure}

However, LLMs face a dual challenge. From the perspective of \textbf{input}, mainstream LLMs encounter limitations when confronted with extensive and structured textual inputs. While it is possible to extend the input length of LLM~\cite{chen2023longlora}, this poses additional challenges and complications. For example, excessively long inputs may hinder the attention mechanism of LLM from effectively encompassing the entire context~\cite{liu2023lost}. Moreover, a significant proportion of real-world texts (\textit{e.g.}, academic papers, social posts) exhibit hierarchical structures rather than strictly adhering to a linear textual order~\cite{zhao2022utilizing,sun2023decoding}. Figure \ref{fig:task} illustrates an exemplary task to identify groundbreaking score of an academic paper. Placing both the paper and its references within a prompt would result in excessive length and compromise the inherent structural relationship. It is a common approach to model hierarchical text information with a tree structure instead of a plain sequence structure. This involves analyzing the relationship between the root and each leaf individually. However, aggregating all leaf information without proper filtering can introduce noise while also being resource-intensive and time-consuming. Therefore, it is crucial to selectively understand and integrate valuable relationships. 

From the perspective of \textbf{output}, while LLMs excel at completing NLP tasks by generating textual responses, practical applications often necessitate directly providing task-required predictions, such as text rating task. While the potential of generative LLMs to improve performance seems promising, existing research indicates a surprising insensitivity to numerical values. A notable example is their inability to accurately compare figures like 9.11 and 9.8. This difficulty arises because LLMs are primarily optimized for discrete text generation rather than precise numerical output, leading to potential inaccuracies and inconsistencies in rating predictions. Despite various methodologies enhancing the generative capabilities of large language models (LLMs), such as parameter-efficient fine-tuning (PEFT) and in-context learning (ICL), challenges in rating tasks requiring continuous numerical predictions remain. While PEFT outperforms ICL in speed and performance in few-shot scenarios \cite{liu2022few}, LLMs still struggle with precise output requirements.

To this end, this study proposes a novel framework, named \textbf{R}ecurrent \textbf{A}lignment with \textbf{H}ard \textbf{A}ttention (\textbf{RAHA}) based on LLMs. Firstly, RAHA employs a frozen LLM to manage message passing within the hierarchical structure of the input. For each pair of root and its respective leaf nodes, the LLM discerns and generates symbolic comparative relationships between them. This paired input preserves the structural information of the root and leaf nodes and is much shorter than putting all leaf texts in one prompt. Here, the evaluation guides the LLM to determine whether a particular leaf requires further scrutiny. This decision functions as the hard attention mechanism, effectively reducing the computational load on the LLM and filtering out irrelevant lower-level details. Then, RAHA leverages another trainable LLM to aggregate all selected symbolic relationships that are considered relevant to the root. This LLM is equipped with a trainable adapter followed by a fully connected layer, enabling it to directly predict text ratings. This targeted aggregation supports more effective prediction. 

Moreover, inspired by the gradual stabilization seen in Markov Chains, we develop a recurrent alignment strategy to enhance task-specific alignment for the trainable LLM. During the training phase, we introduce a special prompt that incorporates the downstream task score predicted by the trainable LLM. Initially, this value is set to \textit{None} and is subsequently updated with the prediction from the previous training iteration. This dynamic updating allows the trainable parameters to progressively learn and refine the alignment from the currently predicted score to the desired target. Furthermore, consistent with this training methodology, during testing, the trainable LLM performs multiple iterative inferences on the same input. This approach ensures that the predictions become increasingly accurate and aligned with the intended outcomes over successive iterations.

We conduct extensive experiments across three hierarchical text rating benchmarks. Our findings demonstrate that the proposed RAHA outperforms existing state-of-the-art methods in predicting task-specific properties. Furthermore, theoretical and empirical analysis highlights its capacity to incrementally approach the most accurate results through iterative inference processes. Finally, we successfully validate the soundness of our approach on other general rating regression datasets.

The main contributions of this study are summarized as follows:
\begin{itemize}
    \item We propose a hard attention mechanism to enable LLMs to effectively and efficiently capture hierarchical relationships, thereby addressing the neglect of content structure in long plain text input.
    \item Drawing inspiration from Markov Chains, we design a recurrent alignment strategy, theoretically and empirically proven to significantly improve the alignment of LLM towards the target value through multiple iterations.
    \item RAHA exhibits superior performance in understanding hierarchical text input to predict rating score, overcoming the limitations of LLMs in continuous numerical tasks.
\end{itemize}

\section{Related Work}
\label{sec:relate}

The essence of human intelligence is characterized by the ability to understand abstract concepts, engage in logical reasoning, and make advanced predictions based on existing knowledge \cite{sternberg1982nature,yu2023nature,huang2022towards}. However, in the era of natural language processing (NLP), despite impressive representation and learning capabilities of neural networks, it is still difficult for them to infer and deduce information from contexts \cite{duan2020machine,wang2022lsat}. This landscape has been dramatically reshaped with the evolution of large language models (LLMs) \cite{brown2020language,workshop2022bloom}, driven by significant upscaling in parameters, data, and computational resources \cite{ouyang2022training,zeng2022glm,touvron2023llama,openai2023gpt}. They exhibit exceptional proficiency for in-context learning (ICL) \cite{brown2020language,min2022rethinking,kojima2022large} across a wide range of NLP tasks \cite{qi2023art,chen2023personalized,wen2023unveiling,du2023task}. One of the key advancements in LLMs is the incorporation of strategies like Chain of Thought (CoT) prompting, which empowers these models to generate reasoning steps and tackle more complex downstream application ~\cite{liu2023pre,wei2022chain,wang2023plan}. 

Notwithstanding the progress made in CoT reasoning \cite{wei2022chain,wang2022self,kojima2022large}, there remains a notable deficiency in current methodologies regarding the processing of hierarchical structures within long text. Numerous studies have focused on identifying and correcting specific thought units where the reasoning process may deviate or require additional information, aiming to produce desired outcomes \cite{yao2023tree,ling2023deductive,yang2023large,wang2023plan}. This prevailing research predominantly concentrates on purely textual content, neglecting the intrinsic hierarchical nature of certain text formats \cite{zhao2022utilizing,sun2023decoding}. In our work, we propose a hard attention mechanism to redress this shortfall by introducing a novel paradigm for enhancing the processing of structured text within CoT reasoning.

The escalation in the scale and adaptability of LLMs has been accompanied by significant advancements in model fine-tuning and adaptation, exemplified by the introduction of various adapter architectures \cite{houlsby2019parameter,pfeiffer2020mad,zaken2021bitfit,hu2022lora}. However, these adaptations have primarily focused on enhancing the model's generation capabilities and have not addressed the limitations of LLMs in directly generating continuous prediction values like text rating. While the prediction of structured continuous numerical values has begun to be explored in some studies \cite{he4685343predicting}, there remains a notable gap in experimentation with large language models in this area.
Concurrently, recent research within LLMs has increasingly focused on recurrent alignment, primarily through prompting techniques and iterative refinement processes \cite{huang2023large,zelikman2022star}. Yet, these methodologies have not sufficiently capitalized on employing the properties from predictive tasks as feedback mechanisms for iterative refinement. Our contribution in this regard is the formulation of a Markov-like recurrent alignment strategy. It represents a novel approach in harnessing the model's output for successive iterative enhancements, thereby augmenting the predictive precision and versatility of LLMs.

\section{Methodology}
\label{sec:model}

\begin{figure*}
\centering
\includegraphics[width=0.99\textwidth]{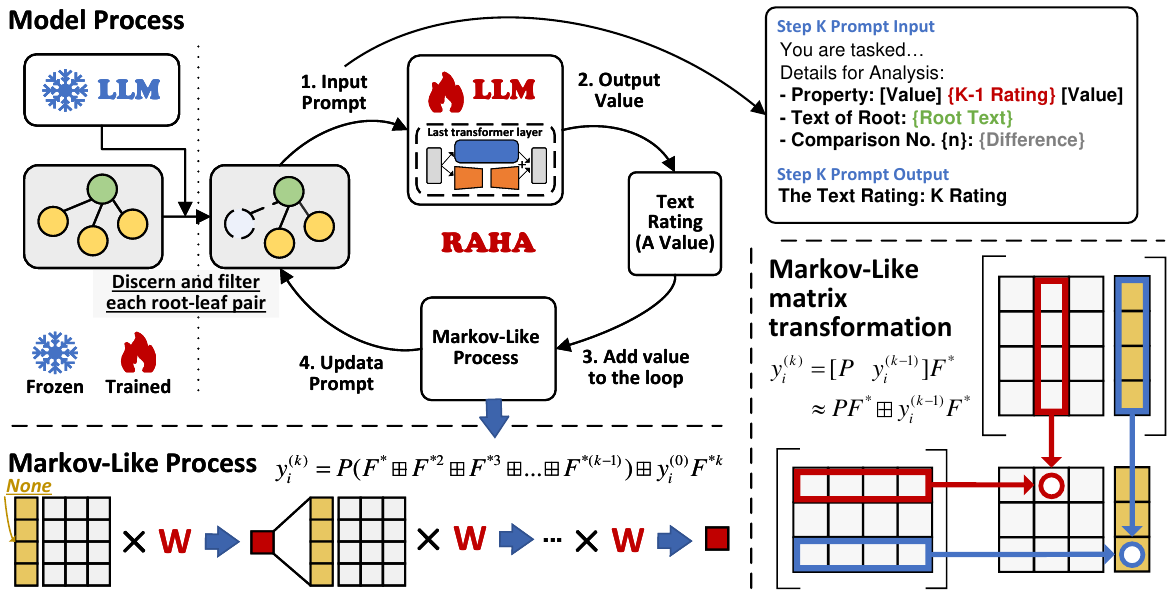}
\caption{The overview of RAHA architecture. A frozen LLM determines connections and generates updates with hard attention scores to filter noise. RAHA incorporates an adapter and fully connected layer within a trainable LLM to predict text rating scores after aggregating updates. During training and testing, the predicted score is fed back into the trainable LLM prompt, refining predictions over multiple iterations.}
\label{fig:model}
\end{figure*}

The proposed framework, RAHA, is depicted in Figure \ref{fig:model}. It includes a tree-based hard attention mechanism that enhances the ability of LLMs to effectively capture hierarchical structures. In addition, a trainable LLM is employed to output hierarchical text rating score. Moreover, we employ a Markov-like recurrent alignment strategy to enable the RAHA to iteratively align with the ground truth of the downstream task.

\subsection{Problem Formulation}
For each sample in our data collection, we represent its hierarchical structure as a tree, which is denote as $\langle r_i, L_i \rangle$. This structure consists of a textual root $r_i$ and a set of $m$ leaves $L_i=\{l_1^{(i)}, l_2^{(i)}, \cdots, l_m^{(i)}\}$. Each leaf $l_j^{(i)}$ serves as the textual root of its own tree and can have its own associated leaves. 

Our framework aims to accomplish an objective with the input $\langle r_i, L_i \rangle$, which is to estimate the text rating $y_i$. By analyzing the hierarchical structure of the data, RAHA can filter meaningful insights and make accurate predictions according to the recurrent alignment strategy.

\subsection{Hard Attention Mechanism}
RAHA framework integrates a tree-based hard attention mechanism to facilitate message passing within a tree structure. It eliminates the necessity for LLMs to grasp the intricate interplay between root and individual leaves within extensive plain texts.

To accomplish this goal, this mechanism firstly utilizes a frozen LLM to figure out the comparative relationship between the root $r_i$ and its $j$-th leaf $l_j^{(i)}$. This process is facilitated by constructing a prompt $p_j^{(i)}$, which contains the following information. Firstly, it provides a clear task description, such as identifying disruptions in papers or predicting potential popularity in social posts. Next, the prompt includes the root text and leaf text along with their respective meta-information. Finally, a well-crafted question is included to extract the necessary features of the root and each leaf that are essential for the task. For a more comprehensive understanding, please refer to the Appendix \ref{appendix:prompt_build_ha} for specific formulation and illustrative examples.

With the provided prompt $p_j^{(i)}$, the LLM can derive two critical pieces of information for each pair of root and child $(r_i, l_j^{(i)})$, which are the hard attention score $a_j^{(i)}$ and a tailored symbolic representation $d_j^{(i)}$:
\begin{equation}
\begin{split}
    p_j^{(i)} & = f_p^{(1)}(r_i, l_j^{(i)}) \\
    a_j^{(i)}, d_j^{(i)} & = \mathcal{F}(p_j^{(i)}) \\
\end{split}
\label{eq:tha}
\end{equation}
where $f_p^{(1)}$ represents the heuristics function for constructing the prompt and $\mathcal{F}$ denotes the frozen LLM.

Here, the hard attention score $a_j^{(i)} \in \{0, 1\}$ is a binary value, that determines whether the leaf $l_j^{(i)}$ deserves further aggregation for the root $r_i$. The symbolic representation $d_j^{(i)}$ serves as an update for the root $r_i$ and provides valuable task-oriented insights. This information captures essential aspects such as the integration, correlation, or distinction between the root $r_i$ and its $j$-th leaf $l_j^{(i)}$. 

Given updates $D_i = [d_1^{(i)}, d_2^{(i)}, \cdots, d_m^{(i)}]$ of the root relative to all leaves, the utilization of hard attention scores $A_i = [a_1^{(i)}, a_2^{(i)}, \cdots, a_m^{(i)}]$ helps filter out potential noise, leading to a reduction in computational consumption:
\begin{equation}
\begin{split}
    D_i^* & = A_i \otimes D_{i}\\
         & = [a_1^{(i)} \otimes d_1^{(i)}, a_2^{(i)} \otimes d_2^{(i)}, \cdots, a_m^{(i)} \otimes d_m^{(i)}]
\end{split}
\label{eq:ha}
\end{equation}
where $\otimes$ denotes the selection operator and $D_i^*$ keeps $m'$ symbolic updates after selection, where $m'\le m$. The valuable updates $D_i^*$ will be aggregated by the subsequent model.

\subsection{Parameter-Efficient Fine-Tuning}
We employ a trainable LLM to complete aggregation of the updates within a tree structure. This LLM is enhanced with Parameter-Efficient Fine-Tuning (PEFT) techniques, which improve its alignment with downstream tasks \cite{houlsby2019parameter}. We integrate trainable parameters $\Delta\bm{W}$ 
as an adapter into the original LLM parameters $\bm{W}_0$ \cite{hu2022lora,liu2022few}. It is represented as:
\begin{equation}
\begin{split}
   & \bm{W}\bm{x} = \bm{W}_0\bm{x} + \Delta\bm{W}\bm{x} = \bm{W}_0\bm{x} + \bm{B}\bm{A}\bm{x} \\
\end{split}
\label{eq:peft}
\end{equation}
where $\bm{B}$ and $\bm{A}$ are both trainable low-rank matrices. In addition, we incorporate a fully connected layer following the hidden representation $\bm{h}$ from the last layer of the LLM. 
\begin{equation}
\begin{split}
   & y = \bm{W}_1\bm{h} \\
\end{split}
\label{eq:fc}
\end{equation}
where the $\bm{W}_1$ is a trainable matrix. This layer facilitates direct prediction of property value for the downstream task. For simplicity, we denote this trainable LLM as $\mathcal{F}^*$.

The prompt for facilitating aggregation of this trainable LLM consists of three key components. Firstly, it includes details about the root $r_i$ of the tree. Secondly, it incorporates the previously filtered updates $D_i^*$. Next, inspired by Markov Chains, it provides the predicted rating score $y_i^*$ of the text required for the task. Finally, we include the task-related question in the prompt. We aim to iteratively bring the predicted value closer to the true value through prior states. It is important to note that at the initial stage, the model has not started the inference yet. As a result, there is no available predicted value, and therefore, this value is set to \textit{None} in the prompt. The prompt can be represented as $p_i$:
\begin{equation}
\begin{split}
    & p_i = f_p^{(2)}([r_i, D_i^*, y_i^*])
\end{split}
\label{eq:prompt}
\end{equation}
where $f_p^{(2)}$ denotes heuristic approach for constructing the prompt $p_i$ and the $y_i^*$ is initialized to \textit{None}, denoted as $\phi$. Please refer to the Appendix \ref{appendix:prompt_build_ra} for specific formulation and illustrative examples. 

\subsection{Recurrent Alignment Strategy}
Many existing studies typically conclude once they complete the previous step. However, we are now considering the possibility of leveraging LLMs to enhance their understanding of inputs based on their previous outputs. Inspired by the principle of Markov Chains, where each state depends on the previous one and converges to a stationary distribution, we propose a recurrent alignment strategy to enhance the learning and inference process of RAHA. Specifically, given the root $r_i$ and filtered updates $D_i^*$, we perform inference multiple times using trainable LLM $\mathcal{F}^*$. The difference of each step is that we update this rating value $y_i^*$ in the prompt function $f_p^{(2)}$ with the model prediction from the previous step. The formulations are shown as follows:
\begin{equation}
\left \{
     \begin{array}{ll}
     y_i^{(1)} & = \mathcal{F}^*(f_p^{(2)}(r_i, D_i^{*}, \phi)) \\
     y_i^{(2)} & = \mathcal{F}^*(f_p^{(2)}(r_i, D_i^{*}, y_i^{(1)})) \\
     & \cdots \\
     y_i^{(k)} & = \mathcal{F}^*(f_p^{(2)}(r_i, D_i^{*}, y_i^{(k-1)})) \\
     \end{array}
\right.
\end{equation}

In this context, each iteration can be viewed as a transition in a Markov Chain, progressively refining the state towards convergence. This strategy offers significant benefits to the model's learning process during the training stage. Since the target output of each iteration is considered the ground truth in the downstream task data, the model gradually approaches the true value based on existing assessments.

During the testing phase, we conduct multiple iterations of the model to perform inference on the same input. This iterative approach allows the model to begin with naive information, advancing step by step towards an accurate hidden representation and progressively aligning itself to the true value. This process is analogous to a Markov Chain reaching its steady-state distribution. Since the model parameters remain unchanged during the testing phase, the process can be considered equivalent to the transition matrix of a Markov Chain. The final predicted value can be expressed as:
\begin{equation}
y_i^{(k)} = P (F^* \boxplus F^{*2} \boxplus F^{*3} \boxplus \cdots \boxplus F^{*(k-1)}) \boxplus y_i^{(0)} F^{*k}
\end{equation}

Generally the spectral radius of the neural network parameter matrix $ F^* $ is less than 1~\cite{blundell2015weight}, so the value can eventually converge to:
\begin{equation}
\lim_{t \to \infty} y_i^{(k)} = P (I - F^{*})^{-1}
\end{equation}
The detailed theoretical proof is in appendix \ref{appendix:formal_proof}. 

\subsection{Training}
Our proposed RAHA integrates two LLMs. The parameters of the first LLM $\mathcal{F}$ remain frozen throughout the process. As for the second LLM $\mathcal{F}^*$, we keep its main parameters $\bm{W}_0$ fixed. We solely employ training data from downstream tasks to optimize its trainable parameters $\Delta\bm{W}$ and $\bm{W}_1$ together, which correspond to the adapter and the fully connected layer, respectively. Specifically, since reasoning $s_i$ has no ground truth, we utilize the property values $y_i$ required by the task to build the mean squared error (MSE) as the objective function:
\begin{equation}
    \mathcal{L} = \frac{1}{2M} \sum_{i=1}^{M} (y_i^{(k)} - y_i)^2
    \label{eq:mse}
\end{equation}
where $M$ is the number of training samples and $y_i^{(k)}$ represent the predicted value for the $i$-the sample in the $k$-th iteration. We conduct a total of $K$ iterations. After each prediction, we will update the prompts for the next iteration. The target value in each round of loss function is the ground truth of the training data. Appendix \ref{appendix:pseudo_code} provides detailed steps for RAHA.

\begin{table*}[htb]
\centering
\resizebox{1.0\linewidth}{!}{
\begin{tabular}{@{}lcccccccc@{}}
\toprule
Model & \multicolumn{2}{c}{DBLP} & \multicolumn{2}{c}{PubMed} & \multicolumn{2}{c}{PatentsView} &\multicolumn{2}{c}{Average}\\ 
\cmidrule(r){2-3} \cmidrule(lr){4-5} \cmidrule(lr){6-7} \cmidrule(l){8-9}
      & MSE & MAE & MSE & MAE & MSE & MAE & MSE & MAE \\
\midrule
SciBERT & 0.072 & 0.119 & \underline{0.025} & 0.116 & 0.069 & 0.121 & 0.055 & 0.119 \\
RoBERTa & 0.061 & 0.094 & 0.030 & \underline{0.112} & 0.069 & 0.100 & 0.053 & \underline{0.102}\\
Bloom-7B & 0.062 & 0.104 & 0.044 & 0.129 & 0.081 & 0.162 & 0.062 & 0.132\\
LLama3 & \underline{0.043} & \underline{0.062} & 0.027 & 0.109 & 0.075 & 0.162 & 0.048 & 0.111 \\
GLM3-6B-32K & 0.045 & 0.091 & 0.056 & 0.182 & \underline{0.042} & \underline{0.088} & \underline{0.047} & 0.120 \\
\midrule
SciBERT-RAHA & 0.043** & 0.077** & 0.038** & 0.119** & 0.060* & 0.104* & 0.047 & 0.100\\
RoBERTa-RAHA & 0.043** & 0.078** & 0.028** & 0.117** & 0.066* & 0.091* & 0.046 & 0.095 \\
Bloom-RAHA & 0.044** & 0.085** & 0.041* & 0.113** & 0.076* & 0.144* & 0.054 & 0.114 \\
LLama3-RAHA & 0.035** & \textbf{0.062**} & 0.025** & 0.109* & 0.045* & 0.090* & 0.035 & 0.087 \\
GLM3-RAHA$_{Forward}$ & \textbf{0.024*} & 0.070** & 0.025* & 0.106** & 0.022* & \textbf{0.084*} & 0.023 & 0.086 \\
GLM3-RAHA$_{Attention}$ & \textbf{0.024*} & 0.078** & \textbf{0.018*} & \textbf{0.072**} & \textbf{0.020*} & 0.099* & \textbf{0.021} & \textbf{0.083}\\
\midrule
\makecell[l]{\hspace{4pt}w/o Hard Attention} & 0.049 & 0.098 & 0.035 & 0.125 & 0.041 & 0.089 & 0.042 & 0.104\\
\makecell[l]{\hspace{4pt}w/o PEFT} & 0.082 & 0.101 & 0.031 & 0.119 & 0.034 & 0.089 & 0.049 & 0.103\\
\makecell[l]{\hspace{4pt}w/o Recurrent Alignment} & \underline{0.025} & \underline{0.085} & \underline{0.028} & \underline{0.110} & \underline{0.023} & \underline{0.085} & \underline{0.025} & \underline{0.093}\\
\bottomrule
\end{tabular}
}
\caption{A comparative results of various language models. The performance is measured in terms of MSE and MAE with lower values indicating better performance. 
The best results are highlighted in \textbf{bold} and \underline{underline} denote the optimal outcomes for each section.
We applied our RAHA framework across all baseline models and examined the effects of PEFT of attention and forward on framework. The ablation studies are based on GLM3-RAHA$_{Forward}$. Notably, the differences observed are statistically significant, as confirmed by a Student's t-test, with an asterisk (*) denoting significant results for the model.}
\label{result}
\end{table*}

\section{Experiments}
\label{sec:exp}

\subsection{Datasets and Evaluation Metrics}
To assess the efficacy of RAHA, we employed five datasets, three of which are hierarchical (DBLP, PubMed, and PatentsView) and two of which are non-hierarchical (ASAP and Splunk). See the Appendix \ref{appendix:data_analysis} for detailed introduction. In the three hierarchical dataset, each is characterized by citation relationships and their respective textual content. Considering the extensive size of these datasets, we randomly select a subset of nearly 10,000 samples from each dataset and allocate 15\% of them for validating and 15\% for testing purposes. 
The primary metric we emphasize is the disruption index \cite{Funk2017ADN,Wu2019LargeTD}, a continuum indicator from -1 to 1 designed to assess the potential of a paper or a patent to transform its respective field.
We use Mean Squared Error (MSE) and Mean Absolute Error (MAE) as the main evaluation metrics. 

\subsection{Baselines} 
We compare RAHA with five baselines. (1) \textbf{SciBERT} \cite{beltagy2019scibert} is a pre-trained language model within the scientific domain. (2) \textbf{RoBERTa} \cite{liu2019roberta} is a robustly optimized BERT. (3) \textbf{BLOOM-7B} \cite{workshop2022bloom} exemplifies advancements in large-scale multi-language processing. (4) \textbf{LLama3} \cite{dubey2024llama} represents the latest iteration in the Llama series of large language models. (5) \textbf{GLM3-6B-32K} \cite{zeng2022glm} is a generative language model based on autoregressive blank Infilling. They're all publicly accessible. For all baselines, we simply add a fully connected layer after their last hidden states for property prediction. Here, we don't compare GPT4 since it lacks the ability to map the input to our numerical target.

\subsection{Experiment Setup}
We implement experiments via PyTorch on a single NVIDIA A800 GPU. Our core experiments, such as ablation test and experiment analysis, are based on GLM3. Optimization of the models is achieved using AdamW optimizer \cite{loshchilov2017decoupled}, with the learning rate set to 1e-5 and the gradient clipping value fixed to 0.2. We set the model to accommodate a maximum input length of 2560. The batch size is set to 4. The low rank of the adapter in the second LLM is 64. We use the PEFT package to insert the adapter in attention or forward part for the last layer of LLM \cite{peft}. The analysis experiment is based on a reasonable analysis of the forward part. The number of training and testing iterations $K$ of RAHA are set to 3 and 5, respectively. The number of epochs is set to 3 for other baselines. The optimal model checkpoint is selected based on performance metrics obtained from the development set.

\subsection{Main Results}
We report the main results on DBLP, PubMed, and PatentView in Table \ref{result}. Overall, we can observe that our framework RAHA achieves the best MSE and MAE in three datasets. LLMs generally outperform PLMs, and the RAHA framework enhances performance across almost all PLMs and LLMs.

The first section of the Table \ref{result} clearly demonstrates that, across the three datasets, the predictive capabilities of large language models generally surpass those of pretrained language models, although some exceptions exist. Notably, within our framework, the incorporation of RAHA consistently results in substantial improvements in the performance of large language models, as well as in the majority of pre-trained language models. Specifically, on the DBLP dataset, RAHA on GLM3 demonstrates superior accuracy, reducing MSE and MAE by 0.021 compared to GLM3. In the PubMed and PatentView datasets, RAHA maintains its leadership, affirming its robustness and adaptability. This improvement underscores RAHA’s precision and consistency in interpreting complex academic metadata. 


The framework's efficacy in these domains can be attributed to its innovative use of a tree-based hard attention mechanism, which methodically navigates through hierarchical data structures, ensuring that significant informational cues are captured and emphasized. Moreover, RAHA’s recurrent alignment strategy enhances its ability to discern and interpret the nuanced linguistic and semantic variations that are critical in fields like biomedical research and patent descriptions.


\begin{figure}[htb]
\centering

\subfloat[DBLP with None]{
\includegraphics[width=0.45\columnwidth]{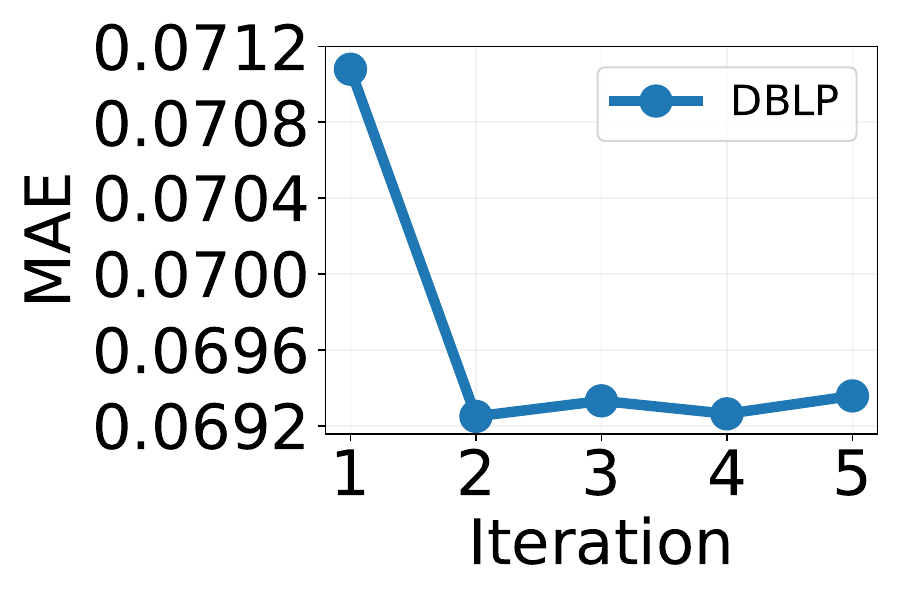}
\label{fig:dblp_mae}
}
\subfloat[DBLP with Random]{
\includegraphics[width=0.45\columnwidth]{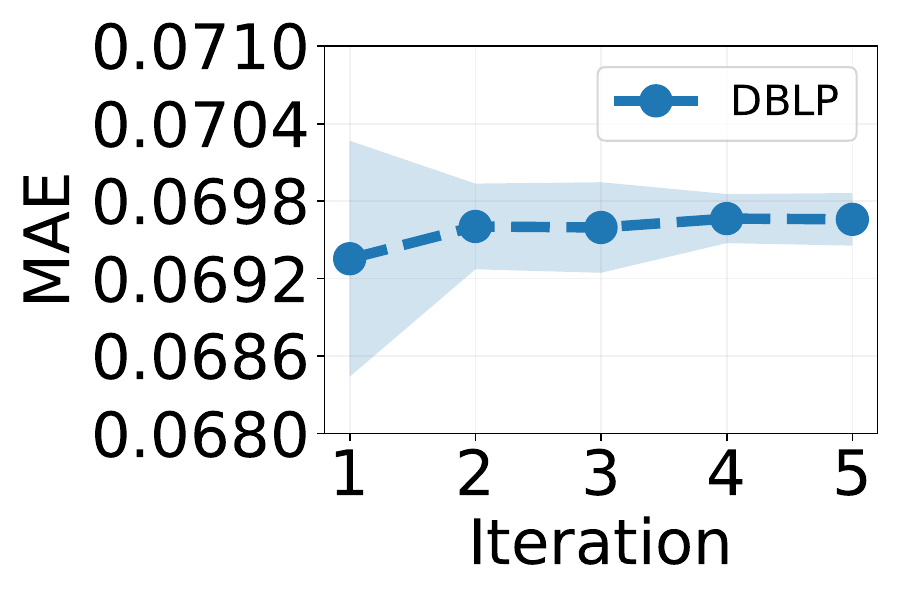}
\label{fig:dblp_random}
}

\subfloat[PMC with None]{
\includegraphics[width=0.45\columnwidth]{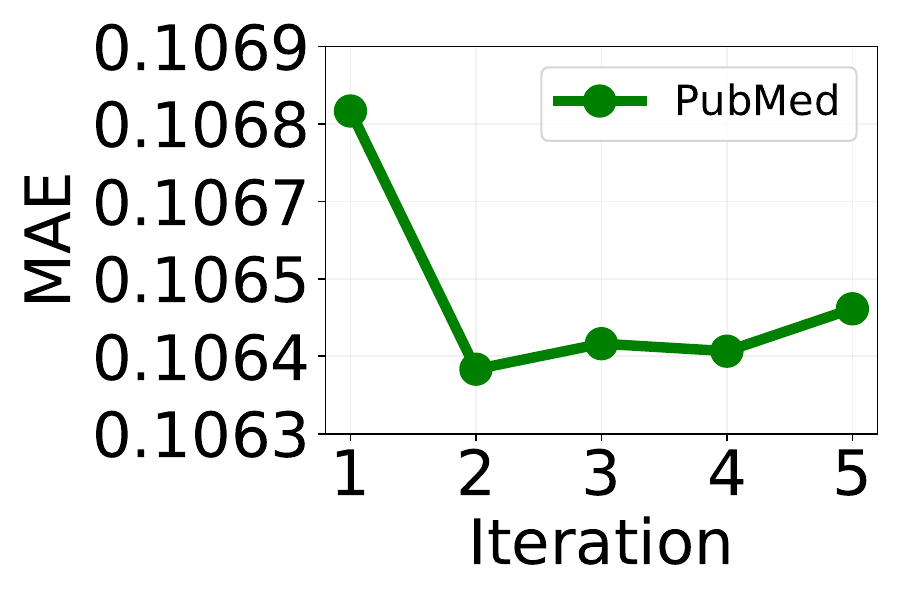} 
\label{fig:pmc_mae}
}
\subfloat[PMC with Random]{
\includegraphics[width=0.45\columnwidth]{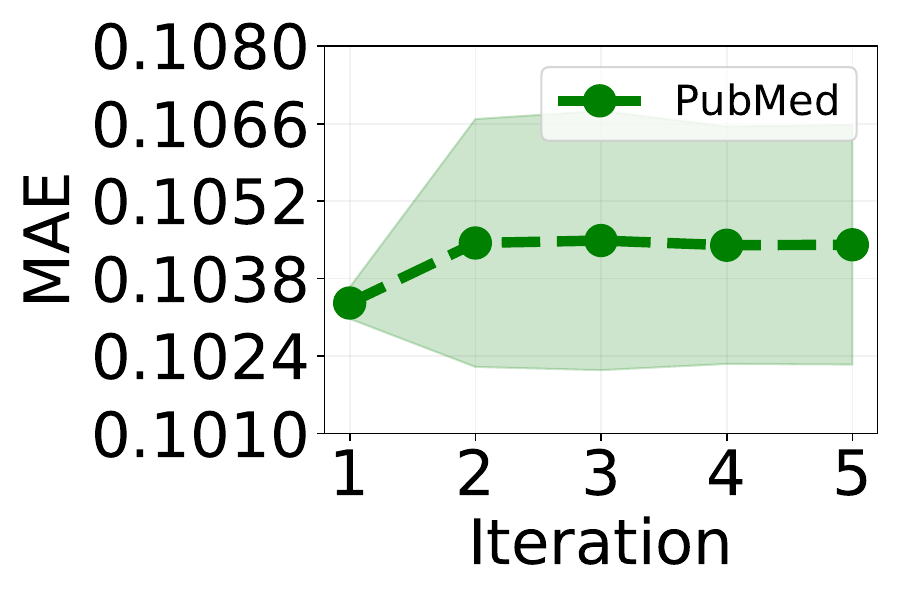} 
\label{fig:pmc_random}
}

\subfloat[Patent with None]{
\includegraphics[width=0.45\columnwidth]{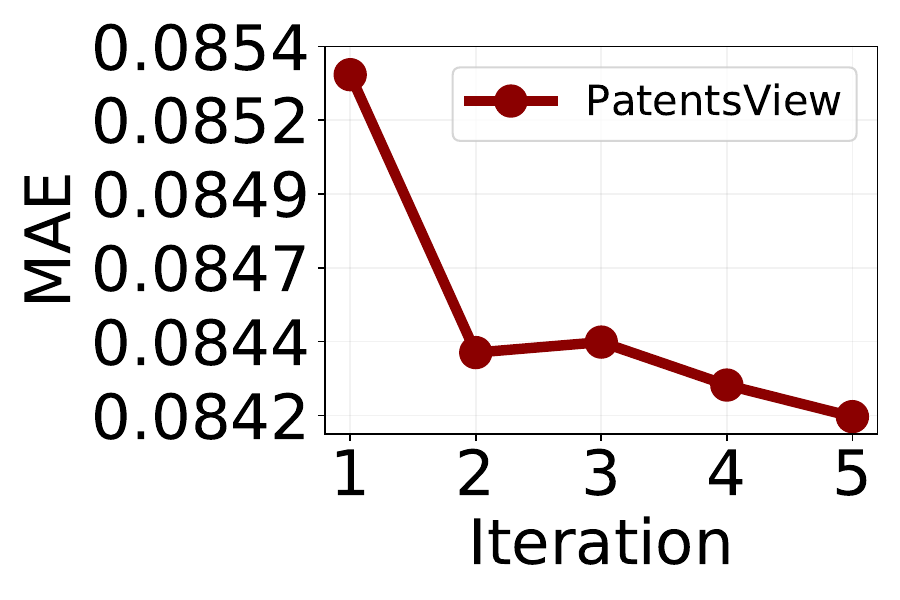} 
\label{fig:patent_mae}
}
\subfloat[Patent with Random]{
\includegraphics[width=0.45\columnwidth]{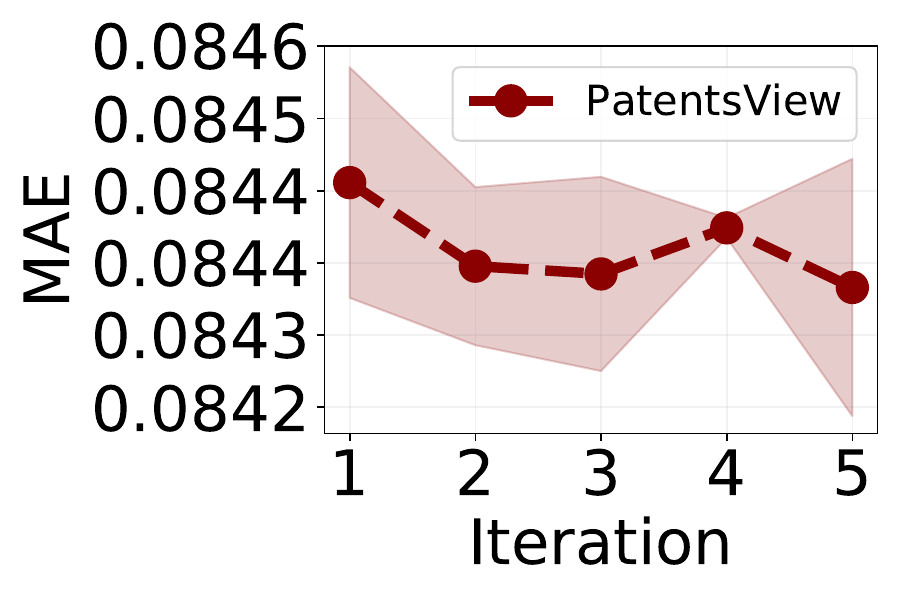} 
\label{fig:patent_random}
}
\caption{Comparison of predictions over multiple iterations during recurrent alignment across three datasets. Figures (a), (c), and (e) show outcomes with the initial prompt set to None. Figures (b), (d), and (f) show results with the initial prompt randomly chosen from -1 to 1.}
\label{fig:comparison}
\end{figure}

\subsection{Ablation Study}
To dissect the contributions of the individual components in our RAHA framework, we conduct ablation studies, as shown in the lower half of Table \ref{result}.

\textbf{(1) RAHA w/o Tree-based hard attention mechanism}: Excluding the hard-attention mechanism leads to a decline in performance across all datasets. This mechanism is crucial for RAHA's ability to process and relate different parts of tree-structured data. Without it, RAHA struggles to pinpoint the most relevant parts of the input text for decision-making, highlighting the importance of understanding the information between the root and leaves.

\textbf{(2) RAHA w/o Parameter-efficient fine-tuning}: Removing the adapter results in the most substantial increases in both MAE and MSE. The adapter enables the second LLM to fine-tune its parameters based on training data. Without it, the second LLM struggles to effectively align with downstream tasks, especially those requiring specific property values, demonstrating the adapter's significance in the architecture.

\textbf{(3) RAHA w/o Recurrent Alignment}: The recurrent alignment strategy iteratively refines outputs based on previous predictions, enhancing the learning process. Without this strategy, there is a slight increase in errors, indicating its critical role in maintaining accuracy and performance by learning from previous predictions.

Furthermore, within the framework of PEFT, we applied LoRA to two distinct components: the attention module and the feed-forward module of the final layer of the transformer. While the performance of LoRA varies across datasets due to its application in different modules, a substantial overall improvement is observed when compared to the baseline model. This suggests that the added modules exhibit a degree of generalizability, as their impact on performance varies across different datasets while still contributing to an overall enhancement in model effectiveness.

\subsection{Predictions over Multiple Iterations}

Figure \ref{fig:comparison} displays the predictions of our RAHA framework over multiple iterations during the test stage. It provides evidence to support our hypothesis that the recurrent alignment strategy allows the fine-tuned LLM to progressively approximate more accurate properties. We use different initialization values in the prompt (see equation \ref{eq:prompt}) to provide broader perspectives for investigating the recurrent alignment strategy. The standard initialization involves using \textit{None} as a value in the prompt. For comparison, we also utilize random initialization for the predicted index, with values ranging from -1 to 1.

As shown in Figure \ref{fig:dblp_mae}, Figure \ref{fig:pmc_mae}, and Figure \ref{fig:patent_mae}, despite fluctuations, the decrease in MAE over gradual iterations demonstrates the ability of RAHA to refine its understanding of the input. This trend suggests that RAHA is not merely fitting to the immediate data but also leveraging its recurrent alignment component to internalize the original input and previous understanding. The ability to improve its performance by iteratively replacing the predicted value in the prompt proves the efficacy of the recurrent alignment strategy.

In contrast, as shown in Figure \ref{fig:pmc_random} and Figure \ref{fig:patent_random}, the result of the recurrent alignment strategy initialized with random values is manifested in a random process according to MAE. The lack of the scratch-to-refinement process we set in place results in models making predictions by guessing rather than reasoning from prior knowledge. This random initialization hampers interpretability  as the predictions are not based on any discernible pattern or learning process. 

Overall, the recurrent alignment strategy is pivotal in aligning RAHA with the downstream task, and predictions cannot be made using unreasonable values from initial randomization. By replacing the predicted value from the previous round to construct the prompt, this approach allows the model to evolve its knowledge in a logical and transparent manner, which is particularly valuable for applications that require reliability and trustworthiness. 

\begin{figure}[ht]
\centering
\subfloat[KL of DBLP]{
\includegraphics[width=0.32\columnwidth]{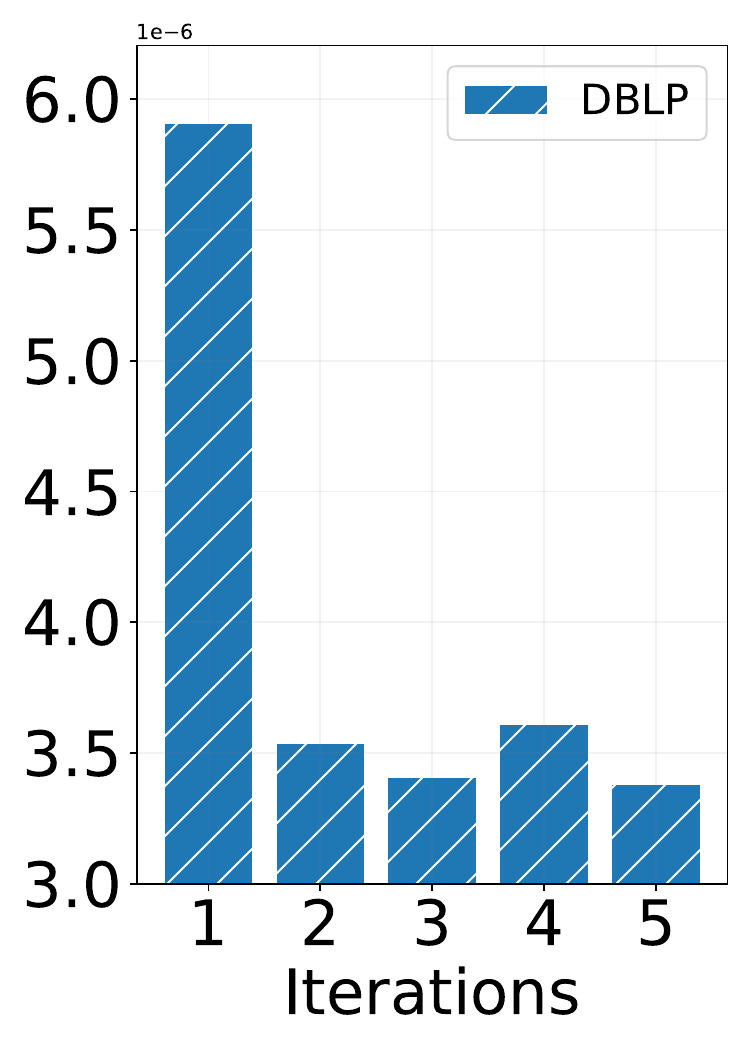}
\label{dblp_kl}
}
\subfloat[KL of PubMed]{
\includegraphics[width=0.32\columnwidth]{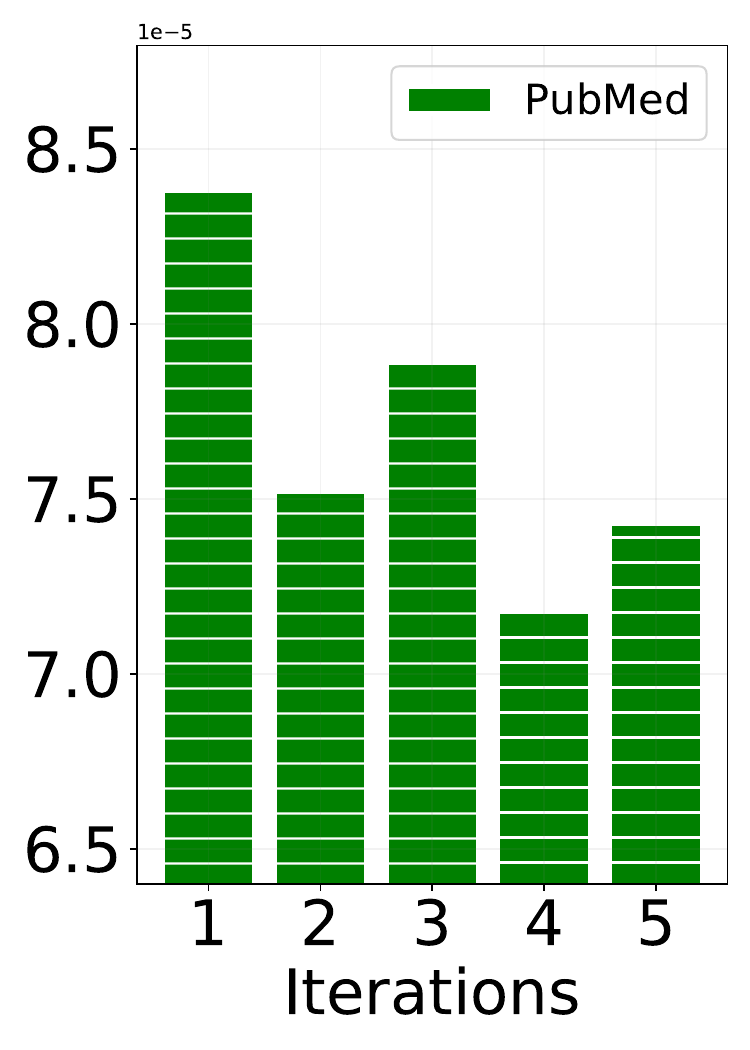} 
\label{pmc_kl}
}
\subfloat[KL of Patents]{
\includegraphics[width=0.32\columnwidth]{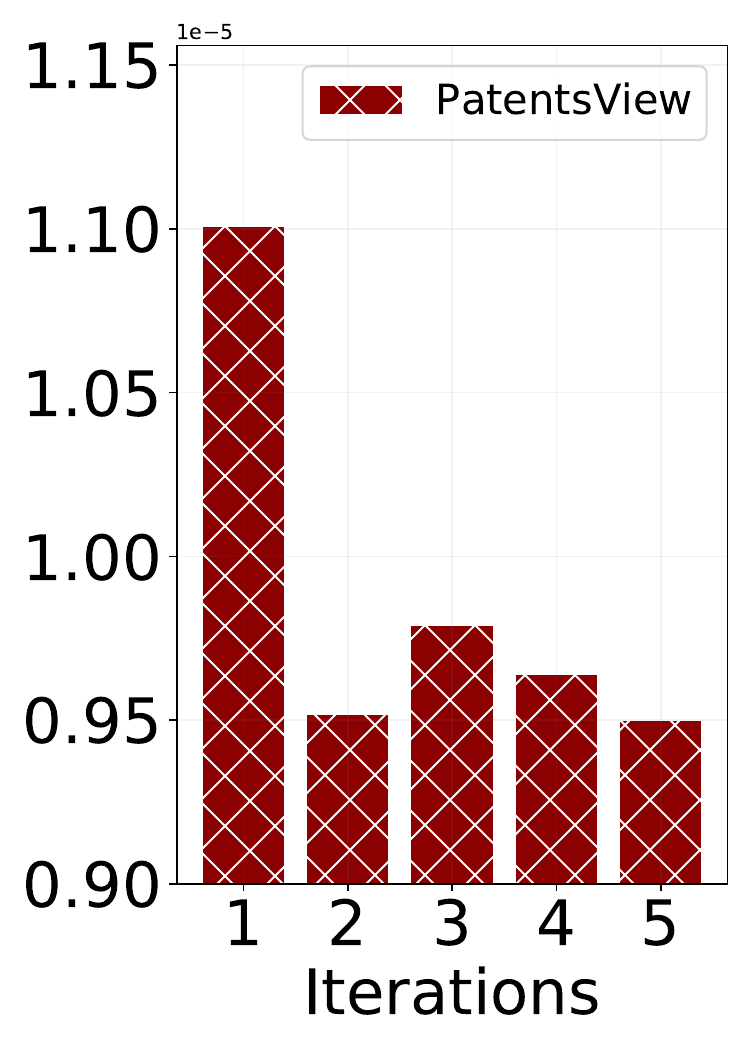} 
\label{patent_kl}
}
\caption{A detailed analysis based on the Kullback-Leibler (KL) divergence over testing iterations across three datasets. It highlights the narrowing gap between the representation of the fine-tuned LLM and the target representation during the recurrent alignment process.}
\label{kl}
\end{figure}

\subsection{Model Representation after Recurrent Alignment}
We provide further insight into the role of the recurrent alignment strategy in driving dynamics of model representation. Since our strategy can enable the trainable LLM to learn the alignment capabilities from scratch to pierce, we assume that directly incorporating the task-desired target truth within the prompt (see equation \ref{eq:prompt}) enables the fine-tuned LLM to derive the target's true representation, facilitating subsequent comparisons with the predicted representation. This simulates a situation where the result obtained through previous understanding is completely correct. We employ the Kullback-Leibler (KL) divergence as a metric to gauge the disparity between the predicted representation extracted by the LLM at each iteration and the target representation. 
Figure \ref{kl} illustrates the trajectories of KL divergence between the target truth and predicted representations over five test iterations across three datasets. Despite occasional fluctuations, the downward trend suggests that RAHA progressively refines its approximation of the target representation. This highlights the effectiveness of the recurrent alignment process. When integrated with the specific predictions from the preceding step, the fine-tuned large language model can better align with downstream tasks by effectively assimilating and aggregating updates. This trend provides a static snapshot of model performance while emphasizing the importance of recurrent alignment iterations.

\subsection{Experiment on Rating Data without Hierarchical Structure}

\begin{table}[htbp]
\centering
\begin{tabular}{@{}lcccc@{}}
\toprule
Model & \multicolumn{2}{c}{ASAP} & \multicolumn{2}{c}{Splunk}\\ 
\cmidrule(r){2-3} \cmidrule(lr){4-5}
      & MSE $\downarrow$ & MAE $\downarrow$ & MSE $\downarrow$ & MAE $\downarrow$\\
\midrule
SciBERT & 0.396 & 0.517 & \underline{\textbf{0.208}} & 0.363\\
Bloom-7b &0.256&  0.446&  0.214&  0.384 \\
GLM3 & \underline{0.252} & \underline{0.439} &  0.214& \underline{0.361}\\
\midrule
RAHA & \textbf{0.249} & \textbf{0.421} & 0.212 & \textbf{0.358}\\
\bottomrule
\end{tabular}
\caption{The performance of various language models on two text rating datasets, ASAP and Splunk, using Mean Squared Error (MSE) and Mean Absolute Error (MAE) as metrics. The best-performing results are emphasized in \textbf{bold}, 
while \underline{underlined} values represent the optimal outcomes within each section.
It is noteworthy that RAHA, built upon GLM3, leverages PEFT in the forward module to achieve these results.}
\label{result1}
\end{table}

To enhance the assessment of the generalization of recurrent alignment, we conduct experiments on two plain text rating datasets. Detailed information of the dataset can be found in Appendix \ref{appendix:data_analysis}.

The Table \ref{result1} presents a performance comparison of various models on these datasets, using MSE and MAE as evaluation metrics. Overall, RAHA demonstrates superior performance across both datasets, particularly excelling in terms of MAE and achieving near-best results in MSE. This highlights RAHA’s robustness and suitability for tasks involving text rating, as well as its ability to effectively capture the nuances in non-hierarchical data. The consistent improvement across these metrics further underscores the significance of the recurrent alignment process in refining model predictions and enhancing task-specific performance.

\section{Conclusion}
\label{sec:conclu}

In this paper, we propose a novel framework called RAHA, that leverages two LLMs to analyze hierarchically structured text. RAHA incorporates a tree-based hard attention mechanism and a recurrent alignment strategy. The tree-based attention enables a frozen LLM to understand the associations between the root and each leaf separately and then selectively choose significant updates for aggregation. This results in a reduction of potential noise in the hierarchical structure and improved utilization of computing resources. The iterative recurrent alignment empowers a trainable LLM to revisit insights gained from previous deliberations, progressively aligning itself with the desired property for downstream tasks. In evaluations on three datasets, RAHA outperforms existing baselines in text rating estimation. Theoretical and empirical analysis reveals that by repeated iterations of prompting the results from the preceding step, RAHA produces hidden representations that gradually approach the optimal representation. This study enhances the abilities of LLMs in handling hierarchical text and aligning with specific tasks.

\section*{Limitation}
We list several limitations in this work that could be improved in the future.

One limitation of our research is the inference time associated with RAHA. The hard attention and iterative recurrent alignment, while beneficial for progressively refining representations, can lead to increased computational overhead. Future efforts should prioritize optimizing the model framework to reduce inference time, enhancing the broader applicability of RAHA.

Additionally, further studies are needed to explore the potential of RAHA in other hierarchical text analysis domains and to validate its performance across a wider range of tasks.

A more rigorous investigation into the principles underlying the recurrent alignment strategy is necessary. Understanding the theoretical foundations and the exact mechanisms through which iterative prompting improves representation alignment can provide deeper insights and guide future enhancements to the model.

\section*{Ethics Statement}
We recognize the ethical implications of our work and the importance of developing and using LLMs responsibly. LLMs are powerful tools that need careful monitoring. While our research aims to improve LLMs, these techniques can also be misused to generate harmful content. We emphasize not placing excessive trust in generated content until LLMs are well-regulated.

\section*{Acknowledgements}
This work is supported by the National Natural Science Foundation of China (72204087, 72104212, 71904058), the Shanghai Planning Office of Philosophy and Social Science Youth Project (2022ETQ001), the "Chen Guang" project supported by Shanghai Municipal Education Commission and Shanghai Education Development Foundation (23CGA28), the Shanghai Pujiang Program (23PJC030), the Natural Science Foundation of Zhejiang Province (LY22G030002), the Fundamental Research Funds for the Central Universities, China, and the 2024 Innovation Evaluation Open Fund, Fudan University (CXPJ2024006). We also appreciate the constructive comments from the anonymous reviewers.

\bibliography{acl_latex}

\appendix

\newpage

\section*{Appendix}
\label{sec:appendix}

\section{Data analysis}
\label{appendix:data_analysis}

In this study, we utilized five diverse datasets to evaluate the performance of our RAHA: DBLP, PubMed, PatentsView, ASAP, and Splunk. Each dataset was split into training, validation, and test sets to ensure robust evaluation and comparison, which is shown as Table \ref{data}.

\textbf{DBLP}: A dataset contains bibliographic information on major computer science journals and proceedings. \url{https://www.aminer.cn/citation}

\textbf{PubMed}: PubMed contains citations and abstracts of biomedical literature from several NLM literature resources, including MEDLINE—the largest component of the PubMed database. \url{https://pubmed.ncbi.nlm.nih.gov/download/}

\textbf{PatentsView}: PatentsView offers publicly accessible patent research data sets with detailed documentation, which focusing on technological and innovation studies. \url{https://patentsview.org/download/data-download-tables}

\textbf{ASAP}: The Automated Student Assessment Prize (ASAP) dataset, sourced from Kaggle, is used for evaluating automated essay scoring systems. \url{https://www.kaggle.com/c/asap-aes/data}

\textbf{Splunk}: A Kaggle competition \textit{Predict WordPress Likes} data, is used for operational intelligence tasks. \url{https://www.kaggle.com/c/predict-wordpress-likes/data}

\begin{table}[htb]
\centering
\begin{tabular}{@{}lcccc@{}}
\toprule
Model & \multicolumn{1}{c}{Train} & \multicolumn{1}{c}{Val} & \multicolumn{1}{c}{Test} & \multicolumn{1}{c}{Total}\\ 
\midrule
DBLP & 6945 & 1488 & 1488 & 9921\\
PubMed & 6956 & 1491 & 1490 & 9937\\
PatentsView & 3988 & 855 & 854 &  5697\\
\midrule
ASAP & 3500 & 750 & 750 &  5000\\
Splunk & 5763 & 1235 & 1235 &  8233\\
\bottomrule
\end{tabular}
\caption{Dataset Splits for RAHA. The table displays the number of instances in the training, validation, and test sets for each dataset (DBLP, PubMed, PatentsView, ASAP, and Splunk).}
\label{data}
\end{table}

\section{Formal Proof of Markov-like Process}
\label{appendix:formal_proof}

In our model, we employ a recurrent alignment strategy, analogous to a Markov chain process, by performing multiple iterations on the same input to refine inference. This approach allows the model to start with naive information and progressively refine towards an accurate representation over time. Given that the model parameters remain unchanged during the testing phase, this iterative process is equivalent to transitions defined by a Markov Chain transition matrix. The mathematical justification proceeds as follows:

\subsection{Definitions}
\begin{itemize}
    \item $y_i^{(k)}$: State of the model at the $k$-th iteration.
    \item $P$: Fixed matrix representation of prompt. 
    \item $F^*$: Represents the fixed parameters of the model during testing, analogous to a transition matrix in a Markov chain.
    \item $\boxplus$: A custom operation defined as follows:
    $ A \boxplus B = (A_1 M + B_1 M) \Vert (A_2 M + B_2 M) $
    Here, $ A $ and $ B $ are matrices that are split into sub-blocks $ A_1, A_2 $ and $ B_1, B_2 $, which are then transformed by matrix $ M $ and recombined.
\end{itemize}

\subsection{Iterative Process Expansion}
The iterative refinement process can be expanded recursively as:
\begin{align*}
y_i^{(k)} &= [P \quad y_i^{(k-1)}] F^* \\
          &= P F^* \boxplus y_i^{(k-1)} F^* \\
          &= P F^* \boxplus (P F^* \boxplus y_i^{(k-2)} F^*) F^* \\
          &= P F^* \boxplus P F^{*2} \boxplus y_i^{(k-2)} F^{*2} \\
          &= \dots \\
          &= P (F^* \boxplus F^{*2} \boxplus \cdots \boxplus F^{*(k-1)}) \boxplus y_i^{(0)} F^{*k} \\
\end{align*}
Define $ S = F^* \boxplus F^{*2} \boxplus \cdots \boxplus F^{*(k-1)} $, where $\boxplus$ operates similarly to addition. 
We can conclude that 
$
\lim_{k \to \infty} S = (I - F^*)^{-1}
$
which implies that
$
y_i^{(k)} \to P (I - F^*)^{-1} \quad \text{as} \quad k \to \infty
$.

The convergence of $ y_i^{(k)} $ to $ P (I - F^*)^{-1} $ as $ k $ approaches infinity can be understood through the lens of stability theory in linear algebra. Since most weights of the neural network are concentrated around zero after training~\cite{blundell2015weight}, the spectral radius of $ F^* $ can be considered to be less than 1. The spectral radius condition, $ \rho(F^*) < 1 $, ensures that the effects of $ F^* $ dampen over successive iterations, leading to the stabilization of $ y_i^{(k)} $. This behavior is analogous to a Markov chain reaching its steady state, where the transition matrix $ F^* $ dictates the evolution of states such that the influence of the initial state progressively wanes, eventually stabilizing at a distribution determined by $ P $ and $ (I - F^*)^{-1} $. This stabilization is crucial in demonstrating that the iterative refinement process under fixed parameters behaves similarly to state transitions in a Markov model, with $ F^* $ serving as a transition-like matrix.

\section{Pseudo Code}
\label{appendix:pseudo_code}

The pseudo-code of our framework is shown in algorithm \ref{alg:raha}.

\begin{algorithm}[tb]
    \caption{RAHA}
    \label{alg:raha}
    \textbf{Input}: hierarchical text $\langle r_i, L_i \rangle$\\
    \textbf{Output}: task-desired property $y_i$

    \begin{algorithmic}[1]
        \WHILE{$1 \leq$ $k$ iteration $\leq K$}
            \FOR{each root and leaf pair $(r_i, s_j^{(i)})$ in $\langle r_i, L_i \rangle$}
                \STATE $p_j^{(i)}$ $\leftarrow$ construct prompt $f_p^{(1)}(r_i, s_j^{(i)})$
                \STATE $a_j^{(i)}, d_j^{(i)}$ $\leftarrow$ conduct inference $\mathcal{F}(p_j^{(i)})$
            \ENDFOR
            \STATE $A_i$ $\leftarrow$ related hard attentions $[a_1^{(i)}, a_2^{(i)}, \cdots, a_m^{(i)}]$    
            \STATE $D_i$ $\leftarrow$ all updates $[d_1^{(i)}, d_2^{(i)}, \cdots, d_m^{(i)}]$ 
            \STATE $D_i^*$ $\leftarrow$ filter out noise $A_i\otimes D_i$
            \IF{$k=1$} 
                \STATE $p_i$ $\leftarrow$ construct aggregation prompt $f_p^{(2)}(r_i, D_i^*, \phi)$
            \ELSE
                \STATE $p_i$ $\leftarrow$ $f_p^{(2)}(r_i, D_i^*, y_i^{(k-1)})$
            \ENDIF
            \STATE $y_i^{(k)}$ $\leftarrow$ conduct inference $\mathcal{F}^*(p_i)$
            \STATE $\mathcal{L}$ $\leftarrow$ compute loss between $y_i^{(k)}$ and $y_i$
            \STATE $\Delta\bm{W}$, $\bm{W}_1$ $\leftarrow$ update parameters via AdamW
        \ENDWHILE
        \STATE \textbf{return} ${y_i^{(k)}}$ 
    \end{algorithmic}
\end{algorithm}

\section{Prompt} 
\label{appendix:prompt_build}

In the appendix section, we present a series of detailed tables that outline the prompts used in the various mechanisms of the RAHA framework. These tables are crucial for understanding the intricacies of how the tree-based hard attention mechanism, parameter-efficient fine-tuning, and recurrent alignment strategy are implemented in practice. Each table provides the structure of prompts used in our experiments, including examples for academic papers and patents. For specific tasks, prompts should be replaced with content that fits the context of the task.

\begin{table}[tb]
    \centering
    \begin{tabular}{p{0.95\columnwidth}}
        \hline
        \textbf{Prompt for Tree-based Hard Attention in Academic Paper Analysis} \\
        \hline
        \textit{Task1}: Determine whether a reference paper is important to a focal paper based on the abstract. Return Import Index is "1" if it is important and "0" if it is not. Don't repeat my inputs, just output the values.\\
        \\
        Example 1:\\
        \textit{Input}:\\
        Focal paper abstract: abstract1\\
        Reference paper abstract: reference1\\
        \textit{Output}: 0\\
        \\
        \textit{Input}:\\
        Focal paper abstract: \{abstract\}\\
        Reference paper abstract: \{reference\}\\
        \textit{Output}:\\
        \hline
        \textit{Task2}: You are now tasked with assessing the disruptive potential in the research area of academic papers. 
Your approach involves contrasting the abstract of a focus paper with the abstracts of its cited references. 
No need to give me abstract's analysis, just output Contrast and Difference.\\
        \\
        Focal paper abstract: \{abstract\}\\
        Reference paper abstract: \{reference\}\\
        \textit{Contrast and Difference}:\\
        \hline
    \end{tabular}
    \caption{Structured Prompts for Tree-Based Hard Attention in Academic Paper Analysis within the RAHA Framework. This table showcases the input format and elucidates how the prompts direct the LLM's focus and analytical processes in handling the hierarchical structures of academic texts.}
    \label{tab:attention}
\end{table}

\subsection{Detailed Prompt for Hard Attention}
\label{appendix:prompt_build_ha}

In the RAHA framework, the integration of a tree-based hard attention mechanism significantly enhances the process of message passing within hierarchical structures. This mechanism streamlines the task for LLMs by reducing the complexity involved in understanding the interplay between the root and individual leaves of a tree within extensive texts. To practically implement this mechanism, we utilize structured prompts that direct the LLM's focus and analytical process. Examples of these structured prompts are illustrated in the following Table \ref{tab:attention}.

In addition to academic papers, the RAHA framework's tree-based hard attention mechanism is adeptly applied to patent analysis. The Table \ref{tab:attention_patent} showcases structured prompts designed for patent analysis.

\begin{table}[tb]
    \centering
    \begin{tabular}{p{0.95\columnwidth}}
        \hline
        \textbf{Prompt for Tree-based Hard Attention in Patent Analysis} \\
        \hline
        \textit{Task1}: Assess the importance of a reference patent based on its abstract in relation to a focal patent. Return an Importance Index as "1" if it is important and "0" if it is not. Do not repeat the inputs, only provide the evaluation.\\
        \\
        Example 1:\\
        \textit{Input}:\\
        Focal Patent abstract: abstract1\\
        Reference Patent abstract: reference1\\
        \textit{Output}: 0\\
        \\
        \textit{Input}:\\
        Focal Patent abstract: \{abstract\}\\
        Reference Patent abstract: \{reference\}\\
        \textit{Output}:\\
        \hline
        \textit{Task2}: You are tasked with analyzing the innovation gap and potential impact between patents. Your job is to contrast the abstract of a focal patent with the abstracts of its related patents. Avoid providing an analysis of the abstracts themselves; focus instead on the contrast and potential differences.
\\
        \\
        Focal Patent abstract: \{abstract\}\\
        Related Patent Abstract: \{reference\}\\
        \textit{Contrast and Difference}:\\
        \hline
    \end{tabular}
    \caption{Structured Prompts for Tree-Based Hard Attention in Patent Analysis within the RAHA Framework. This Table presents examples of how prompts are tailored for assessing the importance and innovation gap between patents, demonstrating the framework's adaptability to different domains.}
    \label{tab:attention_patent}
\end{table}

\begin{table}[t]
    \centering
    \begin{tabular}{p{0.95\columnwidth}}
        \hline
        \textbf{Prompt for Fine-Tuning and recurrent alignment in Academic Paper Analysis} \\
        \hline
        \textit{Task}: You are tasked with assessing the disruptive potential of academic papers. Your primary tool for this analysis is the Disruption Index, a metric ranging from -1 to 1. This index quantifies the level of innovation or breakthrough a paper represents. A higher positive value on the index indicates a significant breakthrough, while negative values suggest a lower level of innovation. \\
        Please provide a detailed analysis based on the contrast and differences between the focus paper and its references. Use the Disruption Index of the focus paper to guide your assessment. Pay special attention to the unique contributions or shortcomings of the focus paper in comparison to the referenced works.
\\
        \\
        \textit{Details for Analysis}: \\
        Determine whether the DINDEX predicted in the previous epoch is high or low: [DINDEX]\{Property\}[DINDEX] \\
        Abstract of Focus Paper: \{abstract\} \\
        Comparison with Reference Paper : \{reference\} \\
        \\
        Based on the above information, analyze the reason for the disruptive nature (or lack thereof) of the focus paper. \\
        \hline
    \end{tabular}
    \caption{Example of a Structured Prompt for Fine-Tuning and recurrent alignment in Academic Paper Analysis within the RAHA Framework. This Table demonstrates how prompts are designed to assess the innovation level of papers using the Disruption Index.}
    \label{tab:self_aca}
\end{table}

\subsection{Detailed Prompt for Fine-Tuning and Recurrent Alignment}
\label{appendix:prompt_build_ra}

In this section, we present a detailed example of a prompt designed specifically for the fine-tuning and recurrent alignment components of the RAHA framework. The {Property} between the [DINDEX] tokens changes iteratively, with the property for this iteration being the output from the previous one. The prompt in Table \ref{tab:self_aca} is tailored for the task of assessing the disruptive potential of academic papers using the Disruption Index. This example illustrates how the prompt structures the analysis process, guiding the model to focus on key indicators and draw meaningful conclusions from the data.

In addition to academic papers, the fine-tuning and recurrent alignment components of the RAHA framework are also effectively applied to the domain of patent analysis. The prompt provided in Table \ref{tab:self_patent} is specifically designed for evaluating the innovation level and potential breakthroughs of patents.

\begin{table}[H]
    \centering
    \begin{tabular}{p{0.95\columnwidth}}
        \hline
        \textbf{Prompt for Fine-Tuning and recurrent alignment in Patent Analysis} \\
        \hline
        \textit{Task}: You are tasked with evaluating the innovation level and potential breakthrough of patents. Your primary tool for this analysis is the Disruption Index, a metric ranging from -1 to 1. This index helps quantify the level of novelty and potential market disruption a patent represents. A higher positive value on the index indicates a significant breakthrough, while negative values suggest incremental or less novel innovations.\\
        Please provide a detailed assessment based on the comparison between the focal patent and its related patents. Consider the Disruption Index of the focal patent to guide your analysis, focusing on the unique contributions or advancements it offers.\\
        \\
        \textit{Details for Analysis}: \\
        Determine whether the DINDEX predicted in the previous epoch is high or low: [DINDEX]\{Property\}[DINDEX] \\
        Abstract of Focus Patent: \{abstract\} \\
        Comparison with Related Patent: \{reference\} \\
        \\
        Based on the above information, predict the Disruption index of the focal patent. \\
        \hline
    \end{tabular}
    \caption{Example of a Structured Prompt for Fine-Tuning and recurrent alignment in Patent Analysis within the RAHA Framework. This Table demonstrates how prompts are designed to assess the innovation level of patents using the Disruption Index.}
    \label{tab:self_patent}
\end{table}

\end{document}